\newcommand{\mystyleflag}{0}

\ifnum\mystyleflag=0
\documentclass[12pt]{article}
\pdfoutput=1
\usepackage{jcappub}
\else
\documentclass[journal,article,submit,pdftex,moreauthors]{Definitions/mdpi} 
\firstpage{1} 
\pubvolume{1}
\issuenum{1}
\articlenumber{0}
\pubyear{2025}
\copyrightyear{2025}
\datereceived{ } 
\daterevised{ } 
\dateaccepted{ } 
\datepublished{ } 
\fi

\usepackage{amsmath}
\usepackage{amssymb}
\usepackage{physics}
\usepackage{slashed}
\usepackage{subfigure}
\usepackage{float}
\usepackage{notoccite}
\usepackage{cancel}
\def\be{\begin{equation}}
\def\ee{\end{equation}}
\def\bea{\begin{eqnarray}}
\def\eea{\end{eqnarray}}

\ifnum\mystyleflag=0
\title{Geometric Learning Dynamics}
\author[1,2]{Vitaly Vanchurin} 
\emailAdd{vitaly.vanchurin@gmail.com}
\affiliation[1]{Artificial Neural Computing, Weston, Florida, 33332, USA}
\affiliation[2]{Duluth Institute for Advanced Study, Duluth, Minnesota, 55804, USA}
\begin{document}
\else
\Title{Geometric Learning Dynamics}
\Author{Vitaly  Vanchurin $^{1,2}$}
\AuthorNames{Vitaly  Vanchurin}
\AuthorCitation{Vanchurin, V.}
\address{%
$^{1}$ \quad Artificial Neural Computing, Weston, Florida, 33332, USA\\
$^{2}$ \quad Duluth Institute for Advanced Study, Duluth, Minnesota, 55804, USA}
\fi

\abstract{We present a unified geometric framework for modeling learning dynamics in physical, biological, and machine learning systems. The theory reveals three fundamental regimes, each emerging from the power-law relationship \( g \propto \kappa^\alpha \) between the metric tensor \( g \) in the space of trainable variables and the noise covariance matrix \( \kappa \). The quantum regime corresponds to \( \alpha = 1 \) and describes Schr\"odinger-like dynamics that emerges from a discrete shift symmetry. The efficient learning regime corresponds to \( \alpha = \tfrac{1}{2} \) and describes very fast machine learning algorithms. The equilibration regime corresponds to \( \alpha = 0 \) and describes classical models of biological evolution. We argue that the emergence of the intermediate regime \( \alpha = \tfrac{1}{2} \) is a key mechanism underlying the emergence of biological complexity.}

\ifnum\mystyleflag=0
\maketitle  
\else
\begin{document}   
\fi

\section{Introduction}\label{sec:intro}

A general learning system can be described as a dynamical system defined by three types of dynamics: boundary, activation, and learning. In classical machine learning, the boundary dynamics corresponds to the dataset, the activation dynamics propagates information through the system via non-trainable variables, and the learning dynamics adjusts the trainable variables to minimize a suitably defined loss function \cite{TTML}. In classical biology, the boundary dynamics corresponds to the environment, the activation dynamics propagates information through an organism, and the learning dynamics adjusts the trainable variables to minimize a suitably defined fitness function \cite{Vanchurin2022MultilevelLearning, Vanchurin2022ThermodynamicsEvolution}. Likewise, in physics, the boundary dynamics corresponds to the specification of boundary conditions, but there is no widely accepted understanding of what (if anything) plays the role of activation or learning dynamics.

To develop such an understanding, it is useful to recognize that the separation of variables into fast and slow degrees of freedom is a familiar concept in physics. In disordered systems, for example, the fast dynamics of individual spins and the slow evolution of coupling constants provide a natural analogy to activation and learning, respectively. This analogy was made explicit in seminal work on neural network models \cite{Hopfield1982}, which established a formal connection between associative memory and spin glasses. In that setting, the neuron activations play the role of fast non-trainable variables, while the synaptic connections act as slow trainable variables that remain quenched on the activation timescale. The geometric approach developed in the present work builds upon this foundation but extends it in two important directions. First, we relax the assumption that fast non-trainable variables are discrete and must reach equilibrium before the slow variables update, allowing for a fully coupled coevolution. Second, we emphasize the role of the emergent metric on the space of trainable variables, a feature absent in standard disordered systems but essential for coordinate independent learning dynamics. This motivates a generally covariant formulation \cite{amari1998natural, guskov2025covariant}, in which the equations transform consistently under reparametrizations of the trainable variables while leaving the underlying dynamics invariant. In doing so, we aim to connect the physics of more general systems than disordered spin glasses with the geometric perspective that has proven valuable in modern machine learning.

If we take this geometric perspective as a starting point, then the question of what constitutes learning and activation in physics becomes not merely analogical but operational. Analytically, it has been argued that the slow trainable variables can represent physical degrees of freedom, such as the values of fields \cite{Vanchurin2024} or the positions of particles \cite{Gusev2025MolecularLearning}, while the fast non-trainable variables can play the role of a complex phase in a quantum description \cite{Katsnelson2021EmergentQuantumness}, or their activation can be described by the dynamics of conjugate variables in classical descriptions \cite{Vanchurin2024}. Numerically, it has been demonstrated that trainable variables appear naturally to represent physical degrees of freedom---such as the positions of nuclei in molecular dynamics simulations---whose learning dynamics converges to Newton's equations of motion in the long-time limit \cite{Gusev2025MolecularLearning}. Moreover, numerous other machine learning based approaches have been successfully applied to molecular systems \cite{Behler2016, Han2018}, often achieving significant speedups over traditional physics-based methods.

The analytical and numerical results can be interpreted in two possible ways, representing the so-called physics-learning duality~\cite{Vanchurin2024, Gusev2025MolecularLearning}. The physics view suggests that the learning dynamics of trainable variables provide a good approximation to the true dynamics of physical degrees of freedom. The opposite, or dual learning view, suggests that it is the learning dynamics that is fundamental, and that the conventional methods of classical and quantum physics merely approximate the true learning behavior. In this article, we adopt the latter, unconventional view: we assume that learning dynamics is more fundamental, from which all artificial and natural phenomena could emerge. In particular, one of our goals is to identify precisely the learning regimes in which quantum physics and biological evolution emerge.

The paper is organized as follows. In Sec.~\ref{sec:ml}, we define a general machine learning system and introduce a geometric framework for modeling learning dynamics. In Sec.~\ref{sec:qphy}, we analyze the quantum regime of the learning dynamics, where the metric tensor is proportional to the noise covariance. In Sec.~\ref{sec:cbio}, we examine biological evolution, which corresponds to the equilibration regime of the learning dynamics in flat space. In Sec.~\ref{sec:nphy}, we introduce geometric learning for a general system and discuss the efficient learning regime. In Sec.~\ref{sec:csys}, we explore possible implications of the geometric learning for complex physical and biological systems. In Sec.~\ref{sec:disc}, we briefly summarize and discuss the main results.

\section{Machine learning}\label{sec:ml}

Consider a machine learning system, such as a neural network~\cite{Galushkin2007, Schmidhuber2015}, defined by three types of dynamics~\cite{TTML}:
\begin{itemize}
    \item \textbf{Boundary dynamics} describes the dataset itself, as well as its coupling to non-trainable boundary neurons---for example, via encoder and decoder maps---which takes place on the time scale \( \tau_b \).

    \item \textbf{Activation dynamics} propagates signals through the network via iterative application of an activation map:
    \begin{equation}
        x_i(t+\tau_a) = f_i(x(t), q(t)), \label{eq:activation}
    \end{equation}
    where \( x_i \) and \( q^\mu \) denote the non-trainable and trainable variables, respectively, and \( \tau_a \) is the time scale of activation dynamics.

    \item \textbf{Learning dynamics} is governed by the covariant gradient descent equation~\cite{amari1998natural, kukleva2024dataset, guskov2025covariant}:
    \begin{equation}
        \tau_l \frac{d q^\mu}{d t} = -\gamma\, g^{\mu\nu}(q)\, \frac{\partial H(q, x)}{\partial q^\nu}, \label{eq:cgd}
    \end{equation}
    where \( \tau_l \) is the time scale of learning dynamics, \( g^{\mu\nu}(q) \) is the inverse metric tensor, and \( H(q,x) \) is the loss function that depends on the trainable variables \( q^\mu \), both explicitly and implicitly through the activation mapping~\eqref{eq:activation}.
\end{itemize}

The time scale of the boundary dynamics, \( \tau_b \), is usually much larger than the time scale of the activation dynamics, \( \tau_b \gg \tau_a \) (e.g., for networks of depth \( L \), \( \tau_b = L \tau_a \)). The time scale of the learning dynamics is the largest of the three: \( \tau_l \gg \tau_b \gg \tau_a \) (e.g., for a mini-batch of size \( M \), \( \tau_l = M \tau_b \)). In this article, we set the time scale of the learning dynamics to \( \tau_l = 1 \), so that the activation and boundary dynamics take place on time scales \( \tau_a \ll \tau_b \ll 1 \).

To describe the learning dynamics as a stochastic process, we model the loss function as:
\begin{equation}
    H(q(t), x(t)) = U(q(t)) + \phi(q(t), t), \label{eq:Uphi}
\end{equation}
where \( U(q) \) is the average loss and \( \phi(q, t) \) is the stochastic component. For starters, we assume that the stochastic fluctuations satisfy the following statistical properties:
\begin{align}
    \langle \phi(q, t) \rangle_{\tau} &= 0, \label{eq:meanvar} \\
    \langle \phi(q, t)\, \phi(q', t') \rangle_{\tau} &= C(q, q')\, \delta(t - t'), \notag
\end{align}
where \( \langle \cdots \rangle_{\tau} \) denotes averaging over a time scale \( \tau \gg \tau_l = 1 \).

Then, the gradient descent equation~\eqref{eq:cgd} can be rewritten as a Langevin equation:
\begin{equation}
    \frac{d q^\mu}{d t} = -\gamma\, g^{\mu\nu}(q)\, \frac{\partial U(q)}{\partial q^\nu} + \eta^\mu(q, t), \label{eq:Langevin}
\end{equation}
where the stochastic noise is defined as
\begin{equation}
    \eta^\mu(q, t) = -\gamma\, g^{\mu\nu}(q)\, \frac{\partial \phi(q, t)}{\partial q^\nu}. \label{eq:noise}
\end{equation}

The corresponding Fokker--Planck equation is given by:
\begin{equation}
    \frac{\partial P}{\partial t} = \gamma \, \frac{1}{\sqrt{\det g}} \, \frac{\partial}{\partial q^\mu} \left( \sqrt{\det g} \, g^{\mu\nu} \left( \frac{\partial U}{\partial q^\nu} \right) P \right) + \frac{\gamma^2}{2} \, \frac{1}{\sqrt{\det g}} \, \frac{\partial}{\partial q^\mu} \left( \sqrt{\det g} \, g^{\mu\alpha} \, \kappa_{\alpha\beta} \, g^{\beta\nu} \left( \frac{\partial P}{\partial q^\nu} \right) \right), \label{eq:FP}
\end{equation}
where the noise covariance matrix is defined as
\begin{equation}
    \kappa_{\alpha\beta}(q) \equiv \left( \frac{\partial}{\partial q^\alpha} \frac{\partial}{\partial q'^\beta} C(q, q') \right)_{q' = q}. \label{eq:kappa}
\end{equation}
Evidently, two key quantities govern the learning dynamics: the metric tensor \( g_{\mu\nu} \) and the noise covariance \( \kappa_{\mu\nu} \). In Ref.~\cite{guskov2025covariant}, it was shown that different machine learning algorithms relate these quantities via a power law:
\begin{equation}
g_{\mu\nu} \propto \left( \kappa^\alpha \right)_{\mu\nu}, \label{eq:a}
\end{equation}
where the exponent takes the values \( \alpha = 1 \) for natural gradient descent~\cite{amari1998natural}, \( \alpha = \tfrac{1}{2} \) for AdaBelief~\cite{zhuang2020adabelief} (an extension of Adam~\cite{kingma2014adam}), and \( \alpha = 0 \) for stochastic gradient descent~\cite{robbins1951stochastic}. In the remainder of the paper, we analyze the significance of these three regimes in the context of physical and biological systems, and explore more general functions \( g(\kappa) \) in relation to complex systems and phase transitions.

\section{Quantum physics}\label{sec:qphy}

For \( g = \kappa \), the Fokker-Planck equation~\eqref{eq:FP} takes a covariant form:
\begin{equation}
    \frac{\partial P}{\partial t} = \gamma \, \frac{1}{\sqrt{\det g}} \, \frac{\partial}{\partial q^\mu} \left( \sqrt{\det g} \, g^{\mu\nu} \left( \frac{\partial U}{\partial q^\nu} \right) P \right) + \frac{\gamma^2}{2} \, \frac{1}{\sqrt{\det g}} \, \frac{\partial}{\partial q^\mu} \left( \sqrt{\det g} \, g^{\mu\nu} \frac{\partial P}{\partial q^\nu} \right), \label{eq:FP1}
\end{equation}
where both the gradient flow and random fluctuations are governed by the same curved geometry, described by the metric tensor \( g_{\mu\nu} \). This limit corresponds to natural gradient descent \cite{amari1998natural} described by \( \alpha = 1 \) in~\eqref{eq:a}. 

The Fokker-Planck equation \eqref{eq:FP1} provides an appropriate description of the learning dynamics with noise given by  Eq.\eqref{eq:meanvar}. More generally,  $\langle \phi(q,t) \rangle_\tau \neq 0$ and the learning dynamics can be described using the principle of stationary entropy production~\cite{Vanchurin2019Entropic}, subject to a constraint on the average change of the loss function:
\begin{equation}
\frac{d(U(q)+\langle \phi(q,t) \rangle_\tau)}{dt} = \frac{\partial \langle \phi(q,t) \rangle_\tau}{\partial t} - \gamma g^{\mu\nu} \frac{\partial (U(q)+\langle \phi(q,t) \rangle_\tau)}{\partial q^\mu} \frac{\partial (U(q)+\langle \phi(q,t) \rangle_\tau)}{\partial q^\nu} = f. \notag
\end{equation}
In learning systems, the dynamics must balance exploration and equilibration. Stationary entropy production offers a natural way to describe this balance, with maximum entropy production corresponding to broad exploration and minimum entropy production to equilibration. We shall discuss this point further in Sec. \ref{sec:nphy} in the context of efficient learning algorithms.

Under the assumption of only `entropic' fluctuations in the sense
\begin{equation}
    g^{\mu\nu} \frac{\partial U}{\partial q^\mu} \frac{\partial \langle \phi \rangle_\tau}{\partial q^\nu} = 0, \label{eq:entropic}
\end{equation}
the constraint becomes
\begin{equation}
    f - \frac{\partial \langle \phi(q,t) \rangle_\tau}{\partial t} + \gamma g^{\mu\nu} \frac{\partial \langle \phi(q,t) \rangle_\tau}{\partial q^\mu} \frac{\partial \langle \phi(q,t) \rangle_\tau}{\partial q^\nu} + \gamma g^{\mu\nu} \frac{\partial U(q)}{\partial q^\mu} \frac{\partial U(q)}{\partial q^\nu} = 0. \label{eq:constraint}
\end{equation}

The Shannon entropy is:
\begin{equation}
S(t) \equiv - \int d^K q \, P(q,t) \log P(q,t)
\end{equation}
and the total entropy production over the time interval \( [0,T] \) is:
\begin{align}
\Delta S &= \int_0^T dt \, \frac{dS}{dt} = -\int_0^T dt \, \int d^K q \, \frac{\partial P}{\partial t} \left( \log P + 1 \right) = -\int_0^T dt \, \int d^K q \, \frac{\partial P}{\partial t} \log P \notag \\
&= \int_0^T dt \, \int d^K q \, \sqrt{\det g} \left( \gamma \, g^{\mu\nu} \frac{\partial U}{\partial q^\nu} \frac{\partial P}{\partial q^\mu} + \frac{\gamma^2}{2} \frac{g^{\mu\nu}}{P} \frac{\partial P}{\partial q^\mu} \frac{\partial P}{\partial q^\nu} \right) \label{eq:production} \\
&= \int_0^T dt \, \int d^K q \, \sqrt{\det g} \left( 2\gamma^2 \frac{\partial \sqrt{P}}{\partial q^\mu} g^{\mu\nu} \frac{\partial \sqrt{P}}{\partial q^\nu} - \gamma \frac{P}{\sqrt{\det g}} \frac{\partial}{\partial q^\mu} \left( \sqrt{\det g} g^{\mu\nu} \frac{\partial U}{\partial q^\nu} \right) \right). \notag
\end{align}
where we used the normalization condition \( \int d^K q\,P = 1 \) or \( \frac{d}{dt}\int d^K q\, P = 0 \), the Fokker-Planck equation \eqref{eq:FP1}, and integrated by parts assuming boundary terms vanish. 

The problem of finding the trajectory corresponding to the stationary entropy production \eqref{eq:production} subject to the constraint \eqref{eq:constraint} can be solved using the method of Lagrange multipliers by defining the action-like functional:
\begin{eqnarray}
\mathcal{S}[P,\phi,\beta] &=& \Delta S + \beta \int_0^T dt\, d^K q \, \sqrt{\det g} \, P \left( f - \frac{\partial \langle\phi\rangle_\tau}{\partial t} + \gamma g^{\mu\nu} \frac{\partial U}{\partial q^\mu} \frac{\partial U}{\partial q^\nu} + \gamma g^{\mu\nu} \frac{\partial \langle\phi\rangle_\tau}{\partial q^\mu} \frac{\partial \langle\phi\rangle_\tau}{\partial q^\nu} \right ) \notag \\
 &=& \int_0^T dt\, d^K q \sqrt{\det g} \bigg[ 2\gamma^2 \frac{\partial \sqrt{P}}{\partial q^\mu} g^{\mu\nu} \frac{\partial \sqrt{P}}{\partial q^\nu} + \beta P \left( V -  \frac{\partial \langle\phi\rangle_\tau}{\partial t} + \gamma g^{\mu\nu} \frac{\partial \langle\phi\rangle_\tau}{\partial q^\mu} \frac{\partial \langle\phi\rangle_\tau}{\partial q^\nu} \right ) \bigg] \notag
\end{eqnarray}
where the effective potential is given by
\begin{equation}
V = \gamma g^{\mu\nu} \frac{\partial U}{\partial q^\mu} \frac{\partial U}{\partial q^\nu} - \frac{\gamma}{\beta} \frac{1}{\sqrt{\det g}} \frac{\partial}{\partial q^\mu} \left( \sqrt{\det g} g^{\mu\nu} \frac{\partial U}{\partial q^\nu} \right) + f. \label{eq:potential}
\end{equation}
Note that for a quadratic average loss function \( U \), the potential \( V \) would also be quadratic, but in general, the two are clearly not the same.

Upon varying with respect to \( \langle\phi\rangle_\tau \) and \( P \) we obtain the Madelung-like equations \cite{Madelung1927}, the continuity equation:
\begin{equation}
\frac{\partial P}{\partial t} - 2 \gamma \, \frac{1}{\sqrt{\det g}} \, \frac{\partial}{\partial q^\mu} \left( \sqrt{\det g} \, P \, g^{\mu\nu} \frac{\partial \langle\phi\rangle_\tau}{\partial q^\nu} \right) = 0, \label{eq:continuity}
\end{equation}
and the quantum Hamilton-Jacobi equation:
\begin{equation}
- \frac{\partial \langle\phi\rangle_\tau}{\partial t} + \gamma \, g^{\mu\nu} \frac{\partial \langle\phi\rangle_\tau}{\partial q^\mu} \frac{\partial \langle\phi\rangle_\tau}{\partial q^\nu} + V + Q = 0, \label{eq:Hamilton-Jacobi}
\end{equation}
where the so-called quantum potential is defined as
\begin{equation}
Q = - \frac{2 \gamma^2}{\beta \sqrt{P}} \, \frac{1}{\sqrt{\det g}} \, \frac{\partial}{\partial q^\mu} \left( \sqrt{\det g} \, g^{\mu\nu} \frac{\partial \sqrt{P}}{\partial q^\nu} \right). \label{eq:qpot}
\end{equation}

Furthermore, if we define a `wavefunction':
\begin{equation}
\psi = \sqrt{P} \exp\left(-\frac{i\langle\phi\rangle_\tau}{\hbar}\right),
\end{equation}
then we obtain a Schr\"odinger-like equation \cite{Schrodinger1926}:
\begin{equation}
i\hbar \frac{\partial \psi}{\partial t} = -\frac{\hbar^2}{2M} \frac{1}{\sqrt{\det g}} \frac{\partial}{\partial q^\mu} \left( \sqrt{\det g} g^{\mu\nu} \frac{\partial \psi}{\partial q^\nu} \right) + V \psi, \label{eq:Schrodinger}
\end{equation}
where the `reduced Planck constant' is
\begin{equation}
\hbar = \sqrt{\frac{2 \gamma}{\beta}},
\end{equation}
and the `mass' is
\begin{equation}
M = \frac{1}{2\gamma}.
\end{equation}
This suggests that in order for $\hbar$ to be constant, the learning rate must be proportional to the Lagrange multiplier \( \gamma \propto \beta \), or the larger the role of the constraint, the larger the learning rate.

Note, however, that unlike the Madelung equations \eqref{eq:continuity} and \eqref{eq:Hamilton-Jacobi}, the Schr\"odinger equation \eqref{eq:Schrodinger} is only valid if a discrete shift in \( \phi \) by Planck constant \( h = 2\pi\hbar \) is an unobservable symmetry \cite{Katsnelson2021EmergentQuantumness}. For example, if \( h \) represents the loss attributed to each neuron, and the total number of neurons \( N \) is unobservable, i.e.
\begin{equation}
\langle\phi\rangle_\tau \cong \langle\phi\rangle_\tau + N h \quad \forall N \in \mathbb{Z}, \label{eq:shift}
\end{equation}
then the Schr\"odinger equation \eqref{eq:Schrodinger} provides a good statistical description of the learning dynamics.

\section{Classical biology}\label{sec:cbio}

In this section we shall examine whether the learning dynamics, and in particular the Fokker--Planck equation \eqref{eq:FP}, can be meaningfully incorporated into classical biological models. This question was already answered affirmatively in the context of the so-called multilevel learning~\cite{Vanchurin2022MultilevelLearning, Vanchurin2022ThermodynamicsEvolution}, while we are here interested in developing a more microscopic description.

Consider biological dynamics (e.g. evolution) in which the state changes with a probability given by the product of the random jump (e.g. genetic drift) and acceptance (e.g. natural selection) probability distributions, i.e.,  
\begin{equation}
P\left(q' \leftarrow q\right) = P_j(q',q) P_a(q',q).
\end{equation}  
We assume that the random jump distribution is a function of the appropriately defined distance between states, i.e.,  
\begin{equation}
P_j(q',q) = P_j(|q' - q|),
\end{equation}  
and the acceptance distribution is such that the detailed balance condition is satisfied:  
\begin{equation}
P\left(q' \leftarrow q\right) P_e\left(q,t\right) = P\left(q \leftarrow q'\right) P_e\left(q',t\right) \label{eq:balance}
\end{equation}  
for the quasi-equilibrium state given by a canonical ensemble:  
\begin{equation}
P_e\left(q,t\right) \propto \exp\left(-\beta H(q,t)\right), \label{eq:equilib}
\end{equation}  
where \(H(q,t)\) is the loss function (e.g. negative log of fitness).  For example, if the acceptance probability is given by the sigmoid function:\footnote{Note that in the related Metropolis algorithm \cite{Metropolis1953} the acceptance probability is given instead by \(P_a\left(q', q\right) = \min(1, \exp\left(\beta H(q,t) - \beta H(q',t)\right))\). }
\begin{equation}
P_a\left(q', q\right) =  \sigma\left(\beta H(q,t) - \beta H(q',t)\right).\label{eq:sigmoid}
\end{equation}  
then the detailed balance condition \eqref{eq:balance} is satisfied: 
\begin{align}
\frac{ P\left(q' \leftarrow q\right)}{P\left(q \leftarrow q'\right)} &= \frac{\sigma\left(\beta H(q,t) - \beta H(q',t)\right)}{\sigma\left(\beta H(q',t) - \beta H(q,t)\right)}\notag \\
&= \frac{1+\exp\left({-\beta H(q',t) + \beta 
H(q,t)}\right)}{1+\exp\left({-\beta H(q,t) + \beta H(q',t)}\right)} \\
&=\exp\left({-\beta H(q',t) + \beta H(q,t)}\right) = \frac{ P_e\left(q',t\right)}{P_e\left(q,t\right)}\notag
\end{align}
where the equilibrium distribution is given by the canonical ensemble \eqref{eq:equilib}.

In the non-equilibrium limit, the expected state at the next time step for the sigmoid acceptance probability distributions~\eqref{eq:sigmoid} is given by
\begin{equation}
\begin{aligned}
\bar{q}^\mu(t+1) = \int d q' P_j(|q' - \bar{q}(t)|) \Big( &\bar{q}^\mu(t) \left[1 - \sigma\left(\beta H(\bar{q}(t),t) - \beta H(q',t)\right)\right] \\
&+ q'^\mu \sigma\left(\beta H(\bar{q}(t),t) - \beta H(q',t)\right) \Big),
\end{aligned}
\end{equation}
or equivalently,
\begin{align}
\frac{d \bar{q}^\mu}{dt}  &=  \int d q'P_j(|q' - \bar{q}(t)|)\left ( \frac{q'^\mu - \bar{q}^\mu(t)}{1 + \exp\left(-\beta H(\bar{q}(t),t) + \beta H(q',t)\right)} \right) \notag \\
&\approx \int d q'P_j(|q' - \bar{q}(t)|) \left( \frac{1}{2} \left( q'^\mu - \bar{q}^\mu(t) \right) - \frac{\beta}{4} \left( q'^\mu - \bar{q}^\mu(t) \right) \left( H(q',t) - H(\bar{q}(t),t) \right) \right) \notag \\
&\approx - \frac{\beta}{4} \int d q'P_j(|q' - \bar{q}(t)|) \left ( \left( q'^\mu - q^\mu(t) \right) \left( q'^\nu - q^\nu(t) \right) \frac{\partial H(\bar{q}(t),t)}{\partial q^\nu} \right ) \notag \\ 
&\approx  - \frac{\beta}{4} C^{\mu\nu} \frac{\partial H(\bar{q}(t),t)}{ \partial q^\nu} \label{eq:averaged}
\end{align}
where the distribution of random jumps is assumed to satisfy
\begin{align}
q^\mu & =  \int dq'\, q'^\mu\, P_j(|q' - q|) , \label{eq:meancov} \\
C^{\mu\nu}& = \int dq'\, (q'^\mu - q^\mu)(q'^\nu - q^\nu)\, P_j(|q' - q|) .  \notag
\end{align}
Eq.~\eqref{eq:averaged} is equivalent to the Lande equation \cite{Lande1976}, which relates the change in mean traits to the product of the genetic covariance matrix and the selection gradient. Note that $\bar{q}^\mu(t)$ denotes the expected trajectory arising from the intrinsic stochasticity of the microscopic model, and does not represent a time-averaged quantity.

Evidently, the model of biological dynamics is equivalent to the covariant gradient descent~\eqref{eq:cgd}:
\begin{equation}
\frac{d \bar{q}^\mu}{dt} = - \frac{\beta}{4} C^{\mu\nu} \frac{\partial H(\bar{q}, t)}{\partial \bar{q}^\nu},
\end{equation}
but since \( C^{\mu\nu} \) does not depend on \( \bar{q} \), it can be globally transformed to a flat form:
\begin{equation}
\frac{\partial \tilde{q}^\mu(\bar{q})}{\partial \bar{q}^\alpha} \frac{\partial \tilde{q}^\nu(\bar{q})}{\partial \bar{q}^\beta} \frac{\beta}{4\gamma} C^{\alpha\beta}(\bar{q}(\tilde{q})) = \tilde{g}^{\mu\nu} = \epsilon \delta^{\mu\nu},
\end{equation}
with an appropriate choice of coordinates \( \tilde{q}^\mu(\bar{q}) \). This is the flat space limit which corresponds to \( \alpha = 0 \) in~\eqref{eq:a}, or to stochastic gradient descent~\cite{robbins1951stochastic}.

On larger time scales, \( \tau \gg \tau_l \), the stochastic dynamics can be described by the Fokker-Planck equation \eqref{eq:FP}:
\begin{equation}
\frac{\partial P}{\partial t} = \frac{\gamma}{\epsilon} \, \frac{\partial}{\partial \tilde{q}^\mu} \left( \delta^{\mu\nu} \left( \frac{\partial U}{\partial \tilde{q}^\nu} \right) P \right) + \frac{\gamma^2}{2} \, \frac{\partial}{\partial \tilde{q}^\mu} \left( \kappa^{\mu\nu} \frac{\partial P}{\partial \tilde{q}^\nu} \right), \label{eq:FP0}
\end{equation}
where the average loss function is decomposed into a time-independent term \( U(\tilde{q}) \) and a stochastic component \( \phi(\tilde{q}, t) \), which satisfies \eqref{eq:meanvar}. Although~\eqref{eq:FP0} describes anisotropic diffusion, extending beyond the typical scope considered in evolutionary biology~\cite{ewens2004mathematical}, it does not capture the system's behavior on the time scales of learning dynamics, \( \tau_l \).

\section{Neural physics}\label{sec:nphy}

To describe a general learning system on the time scales of learning dynamics (i.e., \( \tau_l = 1 \ll \tau \) and \( \tau_l = 1 \gg \tau_b \gg \tau_a \)), we apply the principle of stationary entropy production \cite{Vanchurin2019Entropic} with a constraint on the loss function~\eqref{eq:constraint}, analogous to the approach taken in Sec.~\ref{sec:qphy}. The problem of finding a stationary trajectory can then be solved using the method of Lagrange multipliers by defining the action-like functional:
\begin{align}
\mathcal{S}[P,\phi,\beta] &= \Delta S + \beta \int_0^T dt\, d^K q \, \sqrt{\det g} \, P \left( f - \frac{\partial \langle\phi\rangle_\tau}{\partial t} + \gamma \, g^{\mu\nu} \frac{\partial U}{\partial q^\mu} \frac{\partial U}{\partial q^\nu} + \gamma \, g^{\mu\nu} \frac{\partial \langle\phi\rangle_\tau}{\partial q^\mu} \frac{\partial \langle\phi\rangle_\tau}{\partial q^\nu} \right) \notag \\
&= \int_0^T dt\, d^K q\, \sqrt{\det g} \left[ 2\gamma^2 \frac{\partial \sqrt{P}}{\partial q^\mu} \kappa^{\mu\nu} \frac{\partial \sqrt{P}}{\partial q^\nu} + \beta P \left( V - \frac{\partial \langle\phi\rangle_\tau}{\partial t} + \gamma \, g^{\mu\nu} \frac{\partial \langle\phi\rangle_\tau}{\partial q^\mu} \frac{\partial \langle\phi\rangle_\tau}{\partial q^\nu} \right) \right],
\end{align}
with the effective potential defined in~\eqref{eq:potential}. Variation of the action with respect to \( \langle\phi\rangle_\tau \) yields the continuity equation~\eqref{eq:continuity}, while variation with respect to \( P \) introduces an explicit dependence on the tensor \( \kappa \) into what we shall call the `neural' Hamilton-Jacobi equation:
\begin{equation}
 - \frac{\partial \langle\phi\rangle_\tau}{\partial t} + \gamma \, g^{\mu\nu} \frac{\partial \langle\phi\rangle_\tau}{\partial q^\mu} \frac{\partial \langle\phi\rangle_\tau}{\partial q^\nu} + V + N = 0, \label{eq:neuralHJ}
\end{equation}
where the `neural' potential (a generalization of the quantum potential~\eqref{eq:qpot}) is given by
\begin{equation}
N = - \frac{2 \gamma^2}{\beta \sqrt{P}} \, \frac{1}{\sqrt{\det g}} \, \frac{\partial}{\partial q^\mu} \left( \sqrt{\det g} \, \kappa^{\mu\nu} \frac{\partial \sqrt{P}}{\partial q^\nu} \right). \label{eq:neuralpot}
\end{equation}

For example, in the limit \( g = \sqrt{\kappa} \), which corresponds to the AdaBelief method~\cite{zhuang2020adabelief} (an extension of Adam~\cite{kingma2014adam}), or \( \alpha = \tfrac{1}{2} \) in~\eqref{eq:a}, the learning system can be described by the neural Hamilton-Jacobi equation~\eqref{eq:neuralHJ} with the neural potential:
\begin{equation}
N = - \frac{2 \gamma^2}{\beta \sqrt{P}} \, \frac{1}{\sqrt{\det g}} \, \frac{\partial}{\partial q^\mu} \left( \sqrt{\det g} \, \delta^{\mu\nu} \frac{\partial \sqrt{P}}{\partial q^\nu} \right).
\end{equation}
The covariance of the stochastic noise~\eqref{eq:noise} (in the Langevin equation~\eqref{eq:Langevin}) can be calculated as:
\begin{align}
\langle \eta^\mu \eta^\nu \rangle_{\tau} &= \gamma^2 \, g^{\mu\alpha} g^{\nu\beta} \left\langle \frac{\partial \phi(q)}{\partial q^\alpha} \frac{\partial \phi(q)}{\partial q^\beta} \right\rangle_{\tau} \notag \\
& = \gamma^2 \, g^{\mu\alpha} g^{\nu\beta} \kappa_{\alpha\beta} = \gamma^2 \kappa^{\mu\nu} \\
& = \gamma^2 \delta^{\mu\nu}.\notag 
\label{eq:eomvariance}
\end{align}
In other words, the standard deviation for each trainable variable is exactly the same and is described by the learning rate \( \gamma \). From a learning perspective, this behavior is appropriate when the system is actively exploring the space of all possible solutions. This corresponds to the regime of maximum entropy production, where the system is not yet committed to a particular solution but instead gathers information about the loss landscape. However, once a suitable solution (at least for some trainable variables) is found, the corresponding fluctuations should be suppressed. This corresponds to the regime of minimum entropy production, where the system consolidates its learned state and converges efficiently to a stable equilibrium configuration. 

For instance, if the directional derivatives of the loss function fall below a threshold and the metric is effectively flat, \( g_{\mu\nu} = \epsilon \delta_{\mu\nu} \) or \( \alpha = 0 \) in~\eqref{eq:a}, then
\begin{equation}
\langle \eta^\mu \eta^\nu \rangle_{\tau} = \gamma^2 \, \epsilon^{-2} \delta^{\mu\alpha} \delta^{\nu\beta} \kappa_{\alpha\beta} < \gamma^2 \delta^{\mu\nu}.
\end{equation}
and the neural Hamilton-Jacobi equation \eqref{eq:neuralHJ} is given by 
\begin{equation}
- \frac{\partial \langle\phi\rangle_\tau}{\partial t} + \frac{\gamma}{\epsilon} \, \delta^{\mu\nu} \frac{\partial \langle\phi\rangle_\tau}{\partial q^\mu} \frac{\partial \langle\phi\rangle_\tau}{\partial q^\nu} + V + N = 0, \label{eq:HJ0}
\end{equation}
with neural potential~\eqref{eq:neuralpot}. In this flat-space limit, the dynamics stabilizes as the learning system evolves toward a quasi-stable equilibrium, as observed, for example, in classical biological dynamics (see Sec.~\ref{sec:cbio}).

On longer time scales, \( \tau \gg \tau_l \), the learning dynamics can be described by the Fokker-Planck equation~\eqref{eq:FP}. For example, the regime \( a = 0 \) corresponds to gradient flow in a flat space with stochastic fluctuations in a curved space (assuming \( \det(\kappa) = 1 \)), the regime \( \alpha = \tfrac{1}{2} \) corresponds to gradient flow in a curved space with stochastic fluctuations in a flat space, and the regime \( \alpha = 1 \) corresponds to gradient flow and stochastic fluctuations in the same curved space.

We shall refer to the general models, whether more macroscopic, as described by the Fokker-Planck equation~\eqref{eq:FP}, or more microscopic, as described by the neural Hamilton-Jacobi equations~\eqref{eq:neuralHJ} as neural physics models, to emphasize that the proposed framework is based on neither classical nor quantum physics, although the approximate behavior of both theories emerges in certain limits of the learning dynamics. Note, however, that thus far the significance of the fundamental learning units, or `neurons', has been demonstrated only in the quantum regime \( \alpha = 1 \), characterized by a metric tensor \( g \propto \kappa \); see Sec.~\ref{sec:qphy}.

\section{Complex systems}\label{sec:csys}

In the previous sections, we considered three regimes characterized by the parameter \( \alpha \), which determines how random fluctuations \( \kappa_{\mu\nu} \) influence the metric tensor \( g_{\mu\nu} \) and thereby shape the geometry of the trainable space and its emergent dynamics:
\begin{itemize}
    \item \textbf{Quantum dynamics} (\( \alpha = 1 \)): 
    \begin{equation}
        g \propto \kappa,
    \end{equation}
    
    \item \textbf{Efficient learning} (\( \alpha = \tfrac{1}{2} \)):
    \begin{equation}
        g \propto \sqrt{\kappa},
    \end{equation}

    \item \textbf{Equilibration dynamics} (\( \alpha = 0 \)):
    \begin{equation}
        g \propto I.
    \end{equation}
\end{itemize}
We demonstrated that the \( \alpha = 1 \) regime can describe quantum dynamics of physical systems (Sec.~\ref{sec:qphy}), the \( \alpha = 0 \) regime can describe classical dynamics of biological systems (Sec.~\ref{sec:cbio}), and the \( \alpha = \tfrac{1}{2} \) regime can describe efficient machine learning algorithms (Sec.~\ref{sec:nphy}). Here, we argue that all three regimes are essential for modeling complex physical and biological systems.

To combine the efficient learning and equilibration dynamics regimes, the learning system can employ the metric \cite{guskov2025covariant}:
\begin{equation}
g = \sqrt{\epsilon^2 + \kappa}, \label{eq:12}
\end{equation}
which smoothly interpolates between the regime \( \alpha = \tfrac{1}{2} \) in the large-fluctuation limit and \( \alpha = 0 \) in the small-fluctuation limit, thereby yielding a highly efficient learning algorithm.
However, this formulation does not capture the learning dynamics associated with \( \alpha = 1 \), which, as discussed above, may give rise to quantum-like behavior and potentially offer a quantum advantage in the search for optimal solutions.

A more advanced learning algorithm may therefore need to interpolate between all three regimes, describing fast (\( \alpha = 1 \)), intermediate (\( \alpha = \tfrac{1}{2} \)), and slow (\( \alpha = 0 \)) trainable variables. For example, the system's dynamics could be governed by the metric:
\begin{equation}
g = \sqrt{\epsilon^2 + \kappa + \zeta^2 \kappa^2}. \label{eq:123}
\end{equation}
Fast variables respond rapidly to immediate perturbations, while slow variables evolve gradually, stabilizing the system's long-term behavior. Both types of variables can be found in physical systems (e.g., massless and massive modes) and biological systems (e.g., neutral and core genes). However, intermediate variables arise only in sufficiently complex systems, where they enable adaptation to rapidly changing environments (e.g., adaptive genes).

The emergence of an intermediate scale characterized by \( \alpha = \tfrac{1}{2} \) can be interpreted as a kind of phase transition. For instance, such a transition may be governed by the metric~\eqref{eq:123}, when the parameters evolve from
\begin{equation}
\epsilon \zeta \gg 1, \label{eq:lim1}
\end{equation}
to
\begin{equation}
\epsilon \zeta \ll 1. \label{eq:lim2}
\end{equation}
If \( \lambda_\kappa \) denotes an eigenvalue of the covariance matrix \( \kappa \), then the corresponding eigenvalue of the metric tensor is given by
\begin{equation}
\lambda_g = \sqrt{\epsilon^2 + \lambda_\kappa + \zeta^2 \lambda_\kappa^2}.
\end{equation}
The last two terms become comparable when \( \lambda_\kappa \approx \zeta^{-2} \). If, in this regime, the first term dominates, i.e., \( \epsilon^2 \gg \lambda_\kappa \), then we recover the condition~\eqref{eq:lim1}, corresponding to a system without intermediate variables. Conversely, in the limit \( \epsilon^2 \ll \lambda_\kappa \), the condition~\eqref{eq:lim2} is satisfied, and there emerges a range of scales,
\begin{equation}
\epsilon^2 \lesssim \lambda_\kappa \lesssim \zeta^{-2},\label{eq:range}
\end{equation}
within which efficient learning dynamics can occur along the corresponding directions in the trainable space. Note that in physics, a rigorous demonstration of a phase transition would require taking a thermodynamic limit and carefully tuning the control parameter $\epsilon \zeta$. Although we have not yet performed such an analysis, the observed sudden change in behavior as the learning system switches between regimes $\alpha=0$ and $\alpha=\tfrac{1}{2}$ suggests that a dynamical phase transition may be at play.

It is tempting to interpret this phase transition, from a learning dynamics characterized solely by fast (\( \alpha = 1 \)) and slow (\( \alpha = 0 \)) variables to a dynamics that also incorporates intermediate variables (\( \alpha = \tfrac{1}{2} \)), as the emergence of biological complexity \cite{wolf2018physical} and perhaps as a major transition in evolution \cite{major_transitions, majtranstwo}.
Indeed, a wide range of adaptive scales \eqref{eq:range} may have allowed primitive learning systems to thrive in complex, dynamic environments---laying the foundation for the evolution of complex life forms and for subsequent biological phase transitions \cite{ Vanchurin2022ThermodynamicsEvolution}, as, for example, the phase transitions observed in the evolution of viruses \cite{vanchurin2024quasi}.

In general, geometric learning dynamics is described by the Fokker-Planck equation~\eqref{eq:FP} on time scales $\tau \gg \tau_l$ and by the neural Hamilton-Jacobi equation~\eqref{eq:neuralHJ} on time scales $\tau_l$, with the effective metric in the trainable space $g(\kappa)$ given by the covariance matrix of fluctuations $\kappa$. This implies that the learning system must be capable of storing and retrieving historical information about fluctuations, e.g., in the case of biological evolution, information about past mutations. If such a mechanism can be observationally or experimentally identified, it would provide strong support for the theory of neural physics.

\section{Conclusion}\label{sec:disc}

Our unified framework for geometric learning dynamics reveals fundamental connections between geometry, noise, and emergent behavior across physical, biological and machine learning systems. The key theoretical results that extend beyond traditional modeling approaches can be summarized as follows:
\begin{itemize}
\item On longer time scales \( \tau \gg \tau_l \), the dynamics is described by the Fokker-Planck equation~\eqref{eq:FP}, where the metric tensor \( g_{\mu\nu} \) and the noise covariance matrix \( \kappa_{\mu\nu} \) play essential roles in determining the system's behavior. On shorter time scales characteristic of learning dynamics \( \tau_l \), the dynamics is governed by the neural Hamilton-Jacobi equation~\eqref{eq:neuralHJ}, which emerges from extremizing entropy production subject to a constraint on the loss function.

\item The three fundamental regimes: equilibration dynamics, efficient learning, and quantum dynamics, all emerge from the power-law relationship 
\begin{equation}
g \propto \kappa^\alpha
\end{equation}
between the metric tensor and the noise covariance. The quantum regime (\( \alpha = 1 \)) emerges from the linear relationship and a discrete shift symmetry in the loss function. The efficient learning regime (\( \alpha = \tfrac{1}{2} \)) emerges from the square root relationship and results in a highly efficient machine learning algorithm. The equilibration dynamics regime (\( \alpha = 0 \)) emerges from the flat metric and can describe classical models of biological dynamics.

\item A potentially significant finding is the phase transition in learning complexity that occurs when intermediate variables emerge alongside fast  and slow variables. This transition enables hierarchical adaptation through curvature-noise coupling, a phenomenon that cannot be captured by conventional flat-space evolutionary models. It is tempting to interpret this phase transition as the emergence of biological complexity, or perhaps even the origin of life. However, for geometric learning dynamics to be effective, biological systems must be capable of storing and retrieving historical information about fluctuations. In this sense, biological evolution may not be solely about the survival of the fittest, but also about the survival of the smartest --- those capable of storing and retrieving information about past mutations.

\end{itemize}

Several important research directions emerge from these theoretical findings. In biology, critical next steps might include identifying mechanisms for storing and retrieving information, implementing efficient learning on intermediate scales, and tuning the parameters of the metric tensors. In machine learning, future efforts could focus on developing algorithms that exhibit effectively quantum behavior, efficient adaptive learning dynamics, and equilibration dynamics for solution stabilization. In physics, our framework provides a theoretical basis to study the emergence of quantum field theories and particles~\cite{Vanchurin2024, Gusev2025MolecularLearning}, critical behavior in complex systems~\cite{Katsnelson2023EmergentScale, kukleva2024dataset}, and a possible unification of quantum and gravitational physics within the framework of neural physics~\cite{Vanchurin2, vanchurin2021towards}. We leave the investigation of these and related questions to future research.

{\it Acknowledgments.} The author is grateful to Yaroslav Gusev, Dmitry Guskov, Mikhail Katsnelson, Eugene Koonin and Ekaterina Kukleva for many stimulating discussions.

\bibliographystyle{unsrt}
\bibliography{library}

@article{Vanchurin2,
  author = {Vanchurin, V.},
  title = {The World as a Neural Network
},
  journal = {Entropy},
  volume = {22},
  number = {11},
  pages = {1210},
  year = {2022},
  doi = {10.3390/e22111210},
}

@article{robbins1951stochastic,
  title={A Stochastic Approximation Method},
  author={Robbins, Herbert and Monro, Sutton},
  journal={The Annals of Mathematical Statistics},
  volume={22},
  number={3},
  pages={400--407},
  year={1951},
  publisher={Institute of Mathematical Statistics}
}

@article{kingma2014adam,
  title={Adam: A method for stochastic optimization},
  author={Kingma, Diederik P},
  journal={arXiv preprint arXiv:1412.6980},
  year={2014}
}

@inproceedings{zhuang2020adabelief,
  title={AdaBelief Optimizer: Adapting Stepsizes by the Belief in Observed Gradients},
  author={Zhuang, Juntang and Tang, Tommy and Ding, Yifan and Tatikonda, Sekhar and Dvornek, Nicha C. and Papademetris, Xenophon and Duncan, James S.},
  booktitle={Advances in Neural Information Processing Systems (NeurIPS)},
  year={2020},
  url={https://arxiv.org/abs/2010.07468}
}

@article{guskov2025covariant,
  title={Covariant Gradient Descent},
  author={Guskov, Dmitry and Vanchurin, Vitaly},
  journal={arXiv preprint arXiv:2504.05279},
  year={2025}
}

@article{Vanchurin2024,
  author    = {Vanchurin, Vitaly},
  title     = {Emergent field theories from neural networks},
  journal   = {arXiv preprint arXiv:2411.08138},
  year      = {2024},
  url       = {https://arxiv.org/abs/2411.08138}
}

@article{Vanchurin2022MultilevelLearning,
  title={Toward a theory of evolution as multilevel learning},
  author={Vanchurin, Vitaly and Wolf, Yuri I. and Katsnelson, Mikhail I. and Koonin, Eugene V.},
  journal={Proceedings of the National Academy of Sciences},
  volume={119},
  number={6},
  pages={e2120037119},
  year={2022},
  publisher={National Academy of Sciences},
  doi={10.1073/pnas.2120037119},
  url={https://www.pnas.org/doi/10.1073/pnas.2120037119}
}

@article{Vanchurin2022ThermodynamicsEvolution,
  title={Thermodynamics of evolution and the origin of life},
  author={Vanchurin, Vitaly and Wolf, Yuri I. and Koonin, Eugene V. and Katsnelson, Mikhail I.},
  journal={Proceedings of the National Academy of Sciences},
  volume={119},
  number={6},
  pages={e2120042119},
  year={2022},
  publisher={National Academy of Sciences},
  doi={10.1073/pnas.2120042119},
  url={https://www.pnas.org/doi/10.1073/pnas.2120042119}
}

@article{vanchurin2021towards,
  author  = {Vanchurin, Vitaly},
  title   = {Towards a Theory of Quantum Gravity from Neural Networks},
  journal = {Entropy},
  volume  = {24},
  number  = {1},
  pages   = {7},
  year    = {2021},
  url     = {https://arxiv.org/abs/2111.00903}
}

@article{TTML,
  author =       "V. Vanchurin",
  title =        "Towards a theory of machine learning",
  journal =      "Mach. Learn.: Sci. Technol.",
  volume =       "2",
  number =       {035012},
  pages =        "",
  year =         "2021",
  DOI =          {}
}

@article{Katsnelson2023EmergentScale,
  title = {Emergent scale invariance in neural networks},
  author = {Katsnelson, Mikhail I. and Vanchurin, Vitaly and Westerhout, Tom},
  journal = {Physica A: Statistical Mechanics and its Applications},
  volume = {610},
  pages = {128401},
  year = {2023},
  doi = {10.1016/j.physa.2022.128401},
  url = {https://doi.org/10.1016/j.physa.2022.128401}
}

@article{kukleva2024dataset,
  title={Dataset-learning duality and emergent criticality},
  author={Kukleva, Ekaterina and Vanchurin, Vitaly},
  journal={arXiv preprint arXiv:2405.17391},
  year={2024},
  url={https://arxiv.org/abs/2405.17391},
  doi={10.48550/arXiv.2405.17391}
}

@article{Behler2016,
  author       = {J\"org Behler},
  title        = {Perspective: Machine learning potentials for atomistic simulations},
  journal      = {The Journal of Chemical Physics},
  volume       = {145},
  number       = {17},
  pages        = {170901},
  year         = {2016},
  doi          = {10.1063/1.4966192}
}

@article{Han2018,
  author       = {Jiequn Han and Linfeng Zhang and Roberto Car and Weinan E},
  title        = {Deep Potential: A General Representation of a Many-Body Potential Energy Surface},
  journal      = {Communications in Computational Physics},
  volume       = {23},
  number       = {3},
  pages        = {629--639},
  year         = {2018},
  doi          = {10.4208/CICP.OA-2017-0213}
}

@article{Gusev2025MolecularLearning,
  title={Molecular Learning Dynamics},
  author={Yaroslav Gusev and Vitaly Vanchurin},
  journal={arXiv preprint arXiv:2504.10560},
  year={2025},
  url={https://arxiv.org/abs/2504.10560}
}

@article{Katsnelson2021EmergentQuantumness,
  title={Emergent Quantumness in Neural Networks},
  author={Mikhail I. Katsnelson and Vitaly Vanchurin},
  journal={Foundations of Physics},
  volume={51},
  number={5},
  pages={1--20},
  year={2021},
  doi={10.1007/s10701-021-00503-3},
  url={https://arxiv.org/abs/2012.05082}
}

@book{Galushkin2007,
  author    = {Galushkin, A.I.},
  title     = {Neural Networks Theory},
  publisher = {Springer},
  year      = {2007},
  pages     = {396},
  note      = {ISBN: 978-3-540-48125-6},
  url       = {https://link.springer.com/book/10.1007/978-3-540-48125-6}
}

@article{Schmidhuber2015,
  author    = {Schmidhuber, J.},
  title     = {Deep Learning in Neural Networks: An Overview},
  journal   = {Neural Networks},
  year      = {2015},
  volume    = {61},
  pages     = {85--117},
  doi       = {10.1016/j.neunet.2014.09.003},
  url       = {https://www.sciencedirect.com/science/article/pii/S0893608014002135}
}

@article{amari1998natural,
  title={Natural gradient works efficiently in learning},
  author={Amari, Shun-Ichi},
  journal={Neural computation},
  volume={10},
  number={2},
  pages={251--276},
  year={1998},
  publisher={MIT Press}
}

@article{Vanchurin2019Entropic,
  author  = {Vitaly Vanchurin},
  title   = {Entropic Mechanics: Towards a Stochastic Description of Quantum Mechanics},
  journal = {Foundations of Physics},
  volume  = {50},
  number  = {1},
  pages   = {40--53},
  year    = {2020},
  doi     = {10.1007/s10701-019-00315-6},
  url     = {https://doi.org/10.1007/s10701-019-00315-6}
}

@article{Metropolis1953,
  author    = {Nicholas Metropolis and Arianna W. Rosenbluth and Marshall N. Rosenbluth and Augusta H. Teller and Edward Teller},
  title     = {Equation of State Calculations by Fast Computing Machines},
  journal   = {Journal of Chemical Physics},
  volume    = {21},
  number    = {6},
  pages     = {1087--1092},
  year      = {1953},
  doi       = {10.1063/1.1699114},
  url       = {https://doi.org/10.1063/1.1699114}
}

@article{Madelung1927,
  author = {Erwin Madelung},
  title = {Quantentheorie in hydrodynamischer Form},
  journal = {Zeitschrift f\"ur Physik},
  volume = {40},
  number = {3-4},
  pages = {322?326},
  year = {1927},
  doi = {10.1007/BF01397496}
}

@article{Schrodinger1926,
  author = {Erwin Schr\"odinger},
  title = {Quantisierung als Eigenwertproblem},
  journal = {Annalen der Physik},
  volume = {385},
  number = {13},
  pages = {437?490},
  year = {1926},
  doi = {10.1002/andp.19263851313}
}

@article{majtranstwo,
  author =       "E. Szathmary",
  title =        "Toward major evolutionary transitions theory 2.0.",
  journal =      "PNAS",
  volume =       "112",
  number =       "33",
  pages =        {},
  year =         "2015",
  DOI =          {10.1073/pnas.1421398112}
}

@book{major_transitions,
  title={The Major Transitions in Evolution},
  author={J. M. Smith and E. Szathmary},
  isbn={},
  series={},
  year={1995},
  publisher={Oxford University Press}
}

@book{ewens2004mathematical,
  title     = {Mathematical Population Genetics: I. Theoretical Introduction},
  author    = {Ewens, Warren J.},
  year      = {2004},
  publisher = {Springer},
  isbn      = {978-1-4419-1898-7},
  doi       = {10.1007/978-0-387-21822-9},
  series    = {Interdisciplinary Applied Mathematics},
  volume    = {27},
  edition   = {2nd},
  address   = {New York}
}

@article{vanchurin2024quasi,
  title={Quasi-Equilibrium States and Phase Transitions in Biological Evolution},
  author={Vanchurin, Vitaly and Romanenko, Artem},
  journal={Entropy},
  volume={26},
  number={3},
  pages={201},
  year={2024},
  publisher={MDPI},
  doi={10.3390/e26030201},
  url={https://doi.org/10.3390/e26030201}
}

@article{wolf2018physical,
  title={Physical foundations of biological complexity},
  author={Wolf, Yuri I. and Katsnelson, Mikhail I. and Koonin, Eugene V.},
  journal={Proceedings of the National Academy of Sciences},
  volume={115},
  number={37},
  pages={E8678--E8687},
  year={2018},
  publisher={National Academy of Sciences},
  doi={10.1073/pnas.1807890115},
  url={https://doi.org/10.1073/pnas.1807890115}
}

@article{Hopfield1982,
  title = {Neural networks and physical systems with emergent collective computational abilities},
  author = {Hopfield, John J.},
  journal = {Proceedings of the National Academy of Sciences},
  volume = {79},
  number = {8},
  pages = {2554--2558},
  year = {1982},
  publisher = {National Academy of Sciences},
  doi = {10.1073/pnas.79.8.2554}
}

@article{Lande1976,
  author  = {Lande, Russell},
  title   = {Natural Selection and Random Genetic Drift in Phenotypic Evolution},
  journal = {Evolution},
  year    = {1976},
  volume  = {30},
  number  = {2},
  pages   = {314--334},
  doi     = {10.2307/2407703}
}

\end{document}